%% file: 0_main.tex
\pdfoutput=1

\documentclass[11pt]{article}

\usepackage[preprint]{ACL}

\usepackage{times}
\usepackage{latexsym}
\usepackage{footmisc}

\usepackage[T1]{fontenc}

\usepackage[utf8]{inputenc}

\usepackage{microtype}

\usepackage{inconsolata}

\usepackage{booktabs}
\usepackage{xspace}

\newcommand{\myhl}[1]{{\color{red} \itshape #1}}
\newcommand{\myhlb}[1]{{\itshape\textbf{#1}}}
\newcommand{\ours}{SigExt\xspace}
\newcommand{\gpours}{GP-SigExt\xspace}

\newcommand{\bx}{\mathbf{x}}
\newcommand{\by}{\mathbf{y}}

\usepackage{amsfonts}
\usepackage{amsmath}
\usepackage{amssymb}
\usepackage{graphicx}
\usepackage{enumitem}
\usepackage{listings}
\definecolor{lbcolor}{rgb}{0.95,0.95,0.95}
\title{Salient Information Prompting to Steer Content in\\Prompt-based Abstractive Summarization}

\newcommand*{\affmark}[1][*]{\textsuperscript{#1}}
\newcommand*{\email}[1]{\texttt{#1}}
\renewcommand{\thefootnote}{$\dagger$}

\author{
  Lei Xu\affmark[1],
  Mohammed Asad Karim\affmark[2$^\dagger$],
  Saket Dingliwal\affmark[1],
  Aparna Elangovan\affmark[1] \\
  \affmark[1]Amazon AWS AI Labs\\
  \affmark[2]Carnegie Mellon University\\ 
  \email{\{leixx, skdin, aeg\}@amazon.com} \hspace{2ex} \email{mkarim2@cs.cmu.edu}
}

\begin{document}
\maketitle

\footnotetext{Work done during an internship at AWS AI Labs.}

\renewcommand{\thefootnote}{\arabic{footnote}}

\begin{abstract}
\input{1_abstract}

\end{abstract}

\begin{figure*}[t]
    \centering
    \includegraphics[width=.8\textwidth]{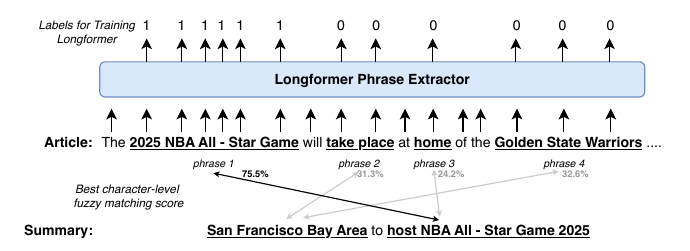}
    \caption{\ours \space-- a finetuned Longformer to extract keyphrases from an article. We construct labels by thresholding the character-level fuzzy matching score between phrases in the article and the summary.}
    \label{fig:model}
\end{figure*}

\input{2_intro}

\input{3_method}

\input{misc_new/tab_main_results}

\begin{figure*}[tb]
\centering
\includegraphics[width=0.9\textwidth]{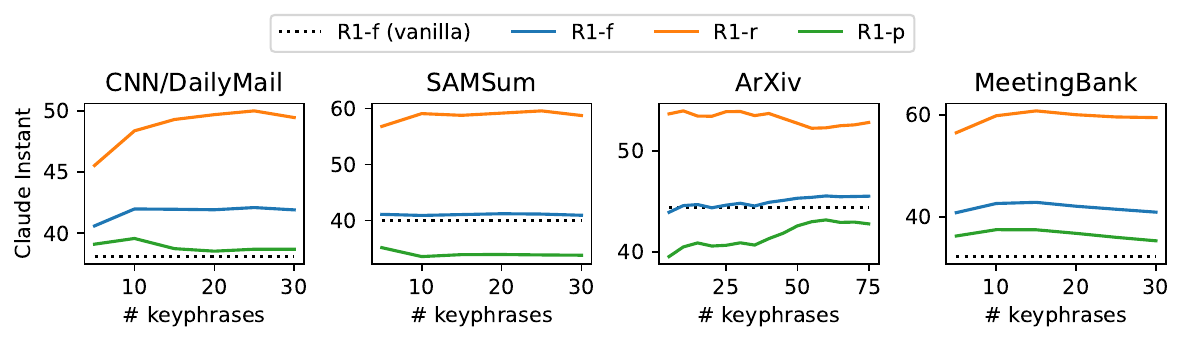}    
\includegraphics[width=0.9\textwidth]{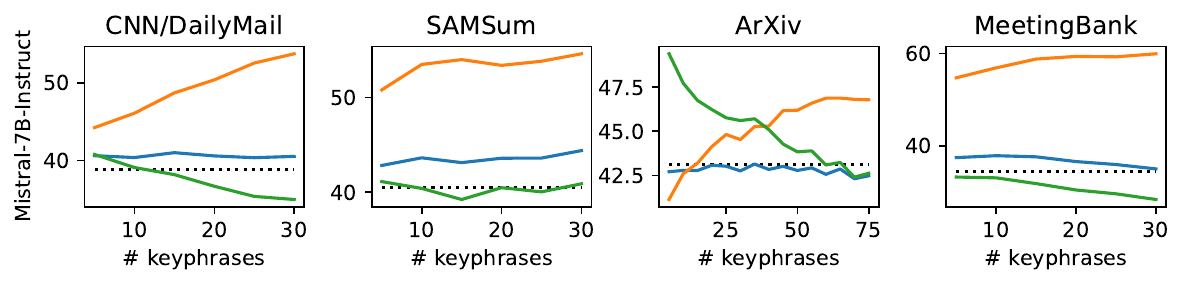}    
\caption{Effect of using different number of keyphrases on the precision-recall trade off.}\label{fig:num_kw}
\end{figure*}

\input{4_exp}

\input{5_con}

\bibliography{custom}

\clearpage
\appendix
\input{6_appendix}

\end{document}

%% file: 1_abstract.tex

Large language models (LLMs) can generate fluent summaries across domains using prompting techniques, reducing the need to train models  for summarization applications. However, crafting effective prompts that guide LLMs to generate summaries with the appropriate level of detail and writing style remains a challenge. In this paper, we explore the use of salient information extracted from the source document to enhance summarization prompts. We show that adding keyphrases in prompts can improve ROUGE F1 and recall, making the generated summaries more similar to the reference and more complete. The number of keyphrases can control the precision-recall trade-off. Furthermore, our analysis reveals that incorporating phrase-level salient information is superior to word- or sentence-level. However, the impact on hallucination is not universally positive across LLMs. To conduct this analysis, we introduce Keyphrase Signal Extractor (\ours), a lightweight model that can be finetuned to extract salient keyphrases. By using \ours, we achieve consistent ROUGE improvements across datasets and open-weight and proprietary LLMs without any LLM customization. Our findings provide insights into leveraging salient information in building prompt-based summarization systems. We release our code at \url{https://github.com/amazon-science/SigExt}

%% file: 2_intro.tex
\section{Introduction}

Abstractive summarization aims to generate concise summaries that capture the most salient information from lengthy source documents. Prior work has shown that emphasizing keywords from source documents can enhance summarization performance on supervised finetuned (SFT) models~\cite{gu-etal-2016-incorporating}. However, existing approaches~\cite{nallapati-etal-2016-abstractive, see-etal-2017-get, liu2021keyword} require extensive modifications to the architecture and loss functions, hindering widespread adoption, especially for large language models (LLMs) with billions of parameters. Recent work~\citep{li2023-stimulus-prompt} trains a separate network using reinforcement learning (RL) to generate keyphrases for LLM prompts, but training RL model is non-trivial due to convergence and stability issues~\citep{Wang2024drlsurvey}. Emphasizing salient information in the prompt can help zero-shot LLMs generate more complete summaries, and steer LLMs to generate summaries that align with the desired use case. However, there is also a lack of analysis on how emphasizing salient information in prompts would affect the LLM behavior.

We first address the challenge of applying salient information to LLMs. We obtain keyphrases using a stand-alone keyphrase signal extractor called \ours, and prompt the LLMs to consider these keyphrases when generating summaries. Unlike prior work relying on complex keyphrase generators optimized for specific LLMs, \ours is LLM-agnostic, allowing leveraging salient information with large API-based models that cannot be finetuned. We demonstrate consistent improvement in ROUGE scores on 4 representative summarization datasets and 3 recent LLMs -- Claude, Mistral~\cite{jiang2023mistral}, and Falcon~\cite{falcon40b} -- highlighting the wide adaptability of our approach. Secondly, we conduct comprehensive experiments using \ours to gain insights into how keyphrases in prompts affect different aspects of summary quality. We show that adding keyphrases improves ROUGE F1 and recall, making the generated summaries more similar to the reference and more complete. Adjusting the number of keyphrases influences the trade-off between precision and recall. Including additional keyphrases in the prompt tends to produce more detailed summaries, enhancing recall. Our findings indicate that using phrase-level salient information is more effective than word- or sentence-level approaches. However, for certain large language models like Mistral, adding keyphrases may lead to more hallucinations. 

Our analysis offers guidance for applying similar strategies in real-world summarization applications. While incorporating salient information is an effective method for enhancing and controlling the completeness of summaries, and using phrase-level granularity proves more effective, the risk of introducing hallucinations must be carefully considered. This risk depends on the specific LLM being used, the method for gathering salient information, and the criticality of the application.

\noindent\textbf{Our key contributions are as follows:} 

\noindent\textbf{1)} We present \ours, a simple yet effective keyphrase extraction model using a finetuned Longformer~\cite{beltagy2020longformer}. Once trained, \ours is LLM-agnostic, enabling performance boost for different LLMs by adding extracted keyphrases in prompts without requiring LLM finetuning.

\noindent\textbf{2)} We provide a comprehensive analysis on the impact of adding salient information in prompts for summarization, including insights on summary length, reference alignment, completeness, and hallucination.

\noindent\textbf{3)} We demonstrate that \ours has cross-domain generalization capability through a general-purpose version (\gpours) pretrained on 7 datasets.

%% file: 3_method.tex
\section{Method}
In this section, we introduce \ours\space-- a keyphrase extractor designed for boosting summarization quality of prompt-based LLMs. Figure~\ref{fig:model} gives an overview. \ours tokenizes the source document into phrases (phrase tokenization is detailed in Section~\ref{sec:phrasetokenisation}), and simultaneously predict whether each phrase is important. To train the model, we create target labels by identifying phrases appear in both the source document and the summary, then optimizing the cross entropy loss. Compared to previous a keyphrase generator that uses RL~\cite{li2023-stimulus-prompt}, \ours allows easier control of keyphrase numbers,  faster training and inference, and better consistency across domains. We directly incorporate keyphrases in prompt, making it generalizable across LLMs. To handle longer input lengths while maintaining efficiency, we build \ours using Longformer, so that training and inference can be done on a single GPU.

\subsection{Phrase tokenization}\label{sec:phrasetokenisation}

Let $\bx = x_1, \ldots, x_n$ be a source document of $n$ tokens, and $\by=y_1,\ldots,y_m$ be the target summary of $m$ tokens. 
The document is segmented into non-overlapping phrases by removing stopwords and puctuation. 
After this, we get a sequence of $T$ non-overlapping phrases, denoted as
$\text{Phrase}(\bx) = [p_i = x_{l_i}\ldots x_{r_i}]_{i=1\ldots T}.$
Similarly, we get $T'$ phrases from the summary denoted as $\text{Phrase}(\by) = [q_i = y_{l_j'}\ldots y_{r_j'}]_{j=1\ldots T'}.$

\subsection{Labels and learning objective}

We label each input phrase by compute the fuzzy matching score 
\[
\resizebox{.7\hsize}{!}{
$\text{fuzz}(a, b)=\frac{|\text{longest\_common\_sequence}(a, b)|}{\max(|a|, |b|)},$}
\]
against all phrases in the summary. If the maximum score exceeds certain threshold $\epsilon$, it is considered a keyphrase, formally 
\[
\resizebox{0.7\hsize}{!}{
$\text{label}(p_i) = 
\begin{cases}
1 & \max_{j\in 1\ldots T'} \text{fuzz}(p_i, q_j) \geq \epsilon, \\
0 & \text{otherwise.}
\end{cases}$
}
\]
We train a classification model to predict the label. Specifically, we use a Longformer and add a classification head on top of each token. We compute the cross entropy loss on tokens that belong to phrases, while ignoring predictions on punctuation and stopword tokens. We apply class balancing weight $\lambda$ when the label of the token is 0.

\subsection{Application of \ours on summarization}

We first finetune \ours on the summarization dataset to get a task-specific keyphrase extractor. 
During inference, we use \ours to extract keyphrases, then wrap the source article with a summarization prompt, and include keyphrases in the prompt. Here is an example prompt:
\begin{lstlisting}
Here is an news article: <text> \nHere are a few keyphrases from the article: <key_phrases> \nPlease write an summary for the article. \nSummary: 
\end{lstlisting}
To select keyphrases, we first score each phrase by calculating the average logits of its tokens. We then select the top-$K$ deduplicated phrases according to their logits scores, removing any duplicates that exceed a fuzzy matching threshold $\epsilon$ and keeping the longer phrase in those cases. We replace \texttt{<key\_phrases>} with comma separated keyphrases. This prompt then serves as the input to the LLM which produces the final summary.

\subsection{Cross domain generalization}\label{sec:cross}
In order to generalize the keyphrase extractor model to new domains without fine-tuning for the target domain, we train a general purpose keyphrase extractor using a combination of 7 datasets.  The datasets are  XSUM~\cite{narayan-etal-2018-dont}, Multi-News~\cite{fabbri-etal-2019-multi}, Gigaword~\cite{nallapati2017summarunner}, Big-Patent~\cite{sharma-etal-2019-bigpatent}, AESLC~\cite{zhang-tetreault-2019-email}, BillSum~\cite{kornilova-eidelman-2019-billsum}, and WikiHow~\cite{koupaee2018wikihow}. We call this general-purpose keyphrase signal extractor model \gpours.

%% file: misc_new/tab_main_results.tex
\begin{table*}[tb]
\centering\small
\begin{tabular}{lccccccccccccc}
\toprule
                  & \multicolumn{3}{c}{\textbf{SAMSum}} & \multicolumn{3}{c}{\textbf{CNN/DailyMail}} & \multicolumn{3}{c}{\textbf{ArXiv}} & \multicolumn{3}{c}{\textbf{MeetingBank}} & \textbf{Avg.} \\
\cmidrule(lr){2-4}\cmidrule(lr){5-7}\cmidrule(lr){8-10}\cmidrule(lr){11-13}\cmidrule(lr){14-14}
\textbf{Method}            & \textbf{R1-f}    & \textbf{RL-f}    & \textbf{R1-r} & \textbf{R1-f}    & \textbf{RL-f}    & \textbf{R1-r} & \textbf{R1-f}    & \textbf{RL-f}    & \textbf{R1-r} & \textbf{R1-f}    & \textbf{RL-f}    & \textbf{R1-r}  & \textbf{$\Delta$R1-f}\\
\midrule
Claude-Ins.    & 40.0    & 30.3    & 52.8   & 38.1      & 23.9      & 41.9      & 44.4    & 23.1   & 53.2   & 32.2      & 21.8     & 43.4    &  \\
+2-stage & 40.3    & \textbf{31.0}    & 46.9   & 39.2      & 24.6      & 48.3      & 44.0    & 22.9   & 50.4   & 30.8      & 20.7     & 43.8    & -0.1 \\
+\gpours        & 40.0    & 30.0    & 57.3   & 40.2      & 24.9      & 47.5      & 44.7    & 23.2   & 53.5   & 36.3      & 25.7     & 53.1    & 1.6 \\
+\ours           & \textbf{41.6}    & 30.9    & \textbf{59.5}   & \textbf{42.0}      & \textbf{26.6}      & \textbf{48.6}      & \textbf{45.2}    & \textbf{23.5}   & \textbf{53.7}   & \textbf{42.3}      & \textbf{31.9}     & \textbf{60.5}    & \textbf{4.1} \\
\midrule
Mistral-7B        & 40.5    & 31.7    & 48.2   & 38.9      & 24.8      & 42.6      & 43.1    & \textbf{24.6}   & 41.6   & 34.4      & 25.2     & 50.3   &  \\
+2-stage & 38.7    & 30.6    & 45.4   & 38.0      & 24.4      & \textbf{48.6}      & 39.5    & 22.0   & 41.9   & 32.0      & 23.5     & 52.0    & -2.2 \\
+\gpours        & 41.9    & 32.2    & 50.7   & 39.5      & 25.2      & 45.3      & 42.8    & 23.8   & 44.7   & 34.1      & 24.7     & 54.8    & 0.4 \\
+\ours           & \textbf{44.1}    & \textbf{33.9}    & \textbf{54.5}   & \textbf{40.9}      & \textbf{26.0}      & 47.9      & \textbf{43.6}    & 24.2   & \textbf{45.2}   & \textbf{37.0}      & \textbf{27.2}     & \textbf{58.7}    & \textbf{2.2} \\
\midrule
Falcon-40B        & 37.1    & 28.7    & 46.3   & 25.7      & 16.4      & 33.8      &         &        &        &           &          &         & \\
+2-stage & 36.1    & 28.1    & 54.1   & \textbf{34.2}      & \textbf{22.1}      & \textbf{53.2}      &         &        &        &           &          &         & 3.8\\
+\gpours        & 38.5    & 29.4    & 54.1   & 31.9      & 20.4      & 42.3      &         &        &        &           &          &         & 3.8 \\
+\ours           & \textbf{39.9}    & \textbf{30.4}    & \textbf{56.1}   & 33.5      & 21.3      & 43.2      &         &        &        &           &          &         & \textbf{5.3} \\
\midrule
0-shot SOTA       & 38.8    & 30.6    & -      & 36.0      & 22.3      & -         & 34.6    & 18.3   &        & 36.4      & 26.8     & -        \\
\bottomrule
\end{tabular}
\caption{Performance of \ours \& \gpours on summarization using Claude Instant, Mistral-7B-Instruct, and Falcon-40B-Instruct. \ours is trained with 1000 examples, while \gpours is not fine-tuned on the dataset. We compare our methods with zero-shot prompting and 2-stage extract-then-abstract baselines. We show ROUGE-1 F-Measure (R1-f), ROUGE-L F-Measure (RL-f), and ROUGE-1 recall (R1-r). The LLMs are not fine-tuned. We directly copy zero-shot SOTA for SAMSum and CNN from \citet{laskar-etal-2023-systematic}, ArXiv from \citet{xiao-etal-2022-primera}, and MeetingBank from \citet{hu-etal-2023-meetingbank}.}\label{tab:zeroshot}
\end{table*}

%% file: 4_exp.tex
\section{Experiments}
\noindent\textbf{Datasets: } 
We select 4 representative datasets -- SAMSum~\cite{gliwa-etal-2019-samsum}, CNN/DailyMail~\cite{nallapati-etal-2016-abstractive}, ArXiv~\cite{cohan-etal-2018-discourse}, and MeetingBank~\cite{hu-etal-2023-meetingbank} -- to evaluate our method. These datasets cover short and long text, as well as regular document and conversation summarization. Dataset details are shown in Table~\ref{tab:dataset} in Appendix. We truncate input text to 4,000 tokens to fit the context window of the Longformer model. We follow the convention to evaluate on 500 randomly sampled examples~\cite{zhang2020pegasus}. We report results averaged on 3 runs.

\noindent\textbf{LLMs and Prompts:} 
We evaluate \ours on Claude-Instant, Mistral-7B-Instruct, and Falcon-40B-Instruct LLMs. We do not use Falcon on ArXiv and MeetingBank datasets due to its limited context window. We manually optimized the prompts for each model and task to achieve competitive zero-shot performance. All prompts are listed in Appendix~\ref{sec:prompt}. 

\noindent\textbf{\ours \& \gpours Parameters:} 
We use Longformer-large (433M) for the keyphrase extractor. We set the fuzzy matching threshold $\epsilon=70\%$, and the class balancing weight $\lambda=0.1$. For \ours, we sample 1000 examples from training set, we train \ours starting with original Longformer-large checkpoint. For \gpours, we sample 10000 examples from each of the 7 dataset mentioned in Sec.~\ref{sec:cross}. 
We train \ours and \gpours for 10 epochs, and use validation set to pick the best checkpoint based on recall@20 (Metric defined in Sec.~\ref{sec:exp-keyword-ext}).

During prompting, we try $K=10, 15, 20$ keyphrases for the CNN, SAMSum, and MeetingBank datasets, and $K=30, 35, 40$ keyphrases for the ArXiv dataset. We pick the best number of keyphrases based on ROUGE scores on the validation set. We also conduct an ablation study on the effect of different numbers of keyphrases.

\noindent\textbf{Baseline:} 
We compare our methods with naive zero-shot prompting. We adapt a \textbf{2-pass} extract-then-abstract method~\cite{zhang-etal-2023-extractive-summarization} to the three LLMs and use it as a baseline. This method uses the LLM to extract sentences from the source document in the first pass, then uses the second pass to revise the extracted sentences into an abstractive summary. We also compare with Directional Stimulus Prompting~\cite{li2023guiding} which utilize reinforcement learning to select good keywords. 

\noindent\textbf{Evaluation Metrics:} 
We compute ROUGE-1/-L F1 scores (abbreviated as \textbf{R1-f}, \textbf{RL-f}) to evaluate summary quality. We also report ROUGE-1 Recall (R1-r) to assess the completeness. We use AlignScore~\cite{zha-etal-2023-alignscore} to evaluate the faithfulness of the summary.

\subsection{Main Results}
Table~\ref{tab:zeroshot} shows the ROUGE scores on all 4 datasets. The F1 scores are improved by using \gpours without any fine-tuning on new datasets. By fine-tuning only the phrase extractor, \ours further improves the score, showing that using a supervisely learned keyphrase extractor can make the LLM generate summaries more similar to the reference. On average, compared to the already strong zero-shot Claude Instant baseline, R1-F improves by 1.6\% with \gpours and 4.1\% with \ours. Similar improvements are also observed on Mistral and Falcon models. Besides F1 scores, adding keyphrases extracted by both \ours and \gpours into the prompts can significantly increase the R1-r score, showing that adding salient information can improve the completeness of the summary. Our method achieves a smaller gain on the ArXiv dataset compared to other datasets. We hypothesize that this is because paper abstracts have a standard format, and the keyphrases they should contain are thus more predictable. As a result, the zero-shot LLM can already identify and include these keyphrases in the output. For other datasets, where the summary is more subjective, our method can help the LLM incorporate proper information in the summary to better align with the reference.

Although the length of the summary slightly increase with the introduction of keyphrases, we do not achieve these improvements by excessively increasing the length of the summary. On average, the length of Claude Instant summaries increases by 4.7 words after adding keyphrases, whereas it increases by 13.6 words for Mistral and 12.3 words for Falcon.

We also compare the performance of \ours with recent Directional Stimulus Prompting baseline on ChatGPT(\texttt{gpt-3.5-turbo}) in Table~\ref{tab:gpt}. We show that \ours can also boost ChatGPT zero-shot performance, and outperform the baseline.

\input{misc/result-gpt}

\subsection{Human Qualitative Check}
To verify the quality of the notes, we follow \citet{liu2023revisiting} and conduct a human evaluation, in which they annotated Atomic Content Units (ACUs) for several public datasets. Each ACU represents a fact that should appear in the summary. We select 50 documents from the CNN and SAMSum datasets, respectively, and ask human annotators to verify whether the given ACU appears in the summary. We report both the \textbf{raw ACU coverage} and \textbf{length-normalized ACU coverage}, as proposed by \citet{liu2023revisiting}. Table~\ref{tab:human} shows that \ours consistently outperforms the vanilla LLM in terms of ACU coverage.
\input{misc_new/tab_human}

\subsection{Number of Keyphrases}
We try different numbers of keyphrases in the prompt for each dataset, and show the ROUGE-1 Precision/Recall/F1 curves in Figure~\ref{fig:num_kw}. The F1 scores of our model are stable when changing the number of keyphrases within a fairly wide range, showing that introducing keyphrases can consistently improve the summary quality.

As we increase the number of keyphrases, there is a clear trend of increasing recall and decreasing precision for the Mistral model. This is less evident for the Claude model. Since we add a length constraint explicitly in the prompt (e.g., "write a summary in 3 sentences"), the Claude model appears to follow these instructions better than the Mistral models. Mistral models tend to try to cover all the keywords provided in the prompt. Consequently, the recall increases significantly when increasing the number of keywords for the Mistral models.

\subsection{Granularity of Salient Information}
We also explore how different granularity of salient information can affect the summarization performance. We compare word-, phrase-, and sentence-level \ours. The results are shown in Table~\ref{tab:granu}. The phrase-level salient information can always achieve top or near-top performance, while the word-level and sentence-level approaches have much larger variance. The word-level information performs poorly on the ArXiv dataset because for academic papers, there are many multi-word phrases that are important in the summary. If these are split, they are no longer helpful for summarization. In contrast, the sentence-level information is not so effective, especially on the MeetingBank dataset. When the dataset is highly abstractive, the important words are dispersed across the document, making it difficult to extract a few sentences to cover the content of the summary (See examples in Appendix Table~\ref{tab:gran_vis}).

\input{misc_new/tab_granularity}

\subsection{Summary Factuality}
As shown in Table~\ref{tab:alignscore}, the effect of adding keyphrases on the AlignScore is LLM and task-specific. For the Claude Instant and Falcon models, the AlignScore is typically improved by incorporating keyphrases. In contrast, the AlignScore always decreases for the Mistral model. These results suggest that keyphrases are not universally helpful for improving the faithfulness of the generated summaries. Table~\ref{tab:mistral-faith} shows a few examples where hallucination is introduced in the summary due to the keyphrases.  The failure pattern is if a keyphrase is negated in the document, Mistral model would ignore the negation. 

\input{misc_new/tab_alignscore}

\subsection{Introducing External Oracle Keyphrases}
We also analyze how external keyphrases which do appear in the source document would affect the performance. We use oracle keyphrases that appears in the reference summary but do not appear in the source document as additional information in the prompt. The ROUGE-1 score and AlignScore are shown on Table~\ref{tab:oracle}. The ROUGE score increases significantly while the AlignScore falls. It indicates that introducing external keyphrases might hurt the factuality of the summary.

\input{misc_new/tab_oracle}

\subsection{Effectiveness of keyphrase extraction}\label{sec:exp-keyword-ext}
In this part, we analyze the effectiveness of the Longformer keyphrase extractor. We define recall@$K$ metric to evaluate the keyphrase extraction performance. We define the recall@$K$ as the recall of \textit{oracle keyphrases} in the top-$K$ deduplicated keywords, where oracle keyphrases are constructed by finding the phrase in the source document with the highest fuzzy match score to each phrase in the target summary. We compare our method with two statistical methods, Rake~\cite{rose2010automatic} and TextRank~\cite{mihalcea-tarau-2004-textrank}. 
Recent work has proposed transformer-based keyphrase extraction models \cite{sun2020sifrank, ding2021attentionrank} that focus on generating noun phrases to better align with human annotation. However, in our setting, the oracle keyphrases are constructed heuristically and are not limited to noun phrases, making these models a poor fit for comparison. Therefore, we do not include them.
The evaluation results are shown on Table~\ref{tab:keyword}. We show that \gpours already outperforms statistical methods. And the fine-tuned \ours achieves additional 5.9\% and 3.7\% improvements on two datasets respectively. 
\input{misc/result-kw-model}

\subsection{Case study}
We show some examples in Appendix Table~\ref{tab:case}. We found the extracted keyphrases can help the LLM incorporate precise details in the summary, hence the summaries better align with the gold summary. In the first two examples, the keyphrases contain exact numbers and times, and the LLM was able to include them in the summary. In the third example, with \ours, the summary covers more topics than the vanilla model. Since we instruct the LLM to ``consider'' these keyphrases, the LLM was able to skip or rephrase some to get more fluent results.

%% file: misc/result-gpt.tex

\begin{table}[htb]
\centering\small
\begin{tabular}{cccccc}
\toprule
\textbf{Method}             & \textbf{\#examples}          & \textbf{R1-f}    & \textbf{RL-f} \\\midrule
Vanilla            & 0             & 38.5     & 25.5   \\
Directional Stimulus    & 4000            & 40.2     & 26.8   \\
\ours              & 4000           & \textbf{42.2}    & \textbf{27.0}   \\\bottomrule
\end{tabular}
\caption{Comparing \ours with baselines using ChatGPT and CNN dataset. }\label{tab:gpt}
\end{table}

%% file: misc_new/tab_human.tex
\begin{table}[htb]
\centering\small
\begin{tabular}{ccccc}
\toprule
       & \multicolumn{2}{c}{\textbf{Raw ACU}} & \multicolumn{2}{c}{\textbf{Nomalized ACU}} \\\cmidrule(lr){2-3}\cmidrule(lr){4-5}
       & \textbf{Claude}   & \textbf{+\ours}  & \textbf{Claude}         & \textbf{+\ours}         \\\midrule
CNN    & 43.8\%           & \textbf{52.4\%}   & 40.7\%                 & \textbf{47.3\%}          \\
SAMSum & 53.6\%           & \textbf{63.3\%}   & 38.4\%                 & \textbf{40.7\%}          \\\bottomrule
\end{tabular}
\caption{ACU coverage human evaluation on CNN and SAMSum using Claude Instant generated summaries.}\label{tab:human}
\end{table}

%% file: misc_new/tab_granularity.tex
\begin{table}[htb]
\centering\small
\begin{tabular}{lcccccc}
\toprule
\textbf{Claude-Instant} & \textbf{R1-f}          & \textbf{RL-f}          & \textbf{R1-f}                & \textbf{RL-f}               \\\midrule
                        & \multicolumn{2}{c}{\textbf{SAMSum}}    & \multicolumn{2}{c}{\textbf{CNN}}        \\
+\ours (word)         & 41.4          & \textbf{30.9} & \textbf{42.0}       & 26.2               \\
+\ours (phrase)       & \textbf{41.6} & \textbf{30.9} & \textbf{42.0}       & \textbf{26.6}      \\
+\ours (sent)         & 39.1          & 29.7          & 40.3                & 25.7               \\\midrule
                        & \multicolumn{2}{c}{\textbf{ArXiv}}     & \multicolumn{2}{c}{\textbf{M.Bank}} \\
+\ours (word)         & 42.2          & 21.0          & 41.9                & 31.7               \\
+\ours (phrase)       & \textbf{45.2} & 23.5          & \textbf{42.3}       & \textbf{31.9}      \\
+\ours (sent)         & 44.8          & \textbf{23.8} & 36.2                & 25.8               \\\bottomrule     
\end{tabular}
\caption{Different granularity of salient information. }\label{tab:granu}
\end{table}

%% file: misc_new/tab_alignscore.tex
\begin{table}[htb]
\centering\small
\begin{tabular}{lcccc}
\toprule
               & \textbf{SamSum} & \textbf{CNN}  & \textbf{ArXiv} & \textbf{M.Bank} \\\midrule
Claude Ins. & 85.8   & \textbf{83.8} & 53.7  & 73.1        \\
+\ours       & \textbf{88.0}   & 82.3 & \textbf{60.0}  & \textbf{74.7}        \\
\midrule
Mistral-7B & \textbf{88.9} & \textbf{88.8} & \textbf{56.9} & \textbf{79.1} \\
+\ours  & 84.7 & 87.0 & 49.5 & 77.1 \\
\midrule
Falcon-40B & \textbf{81.6} & 67.7 & & \\
+\ours & \textbf{81.6} & \textbf{75.0} & & \\
\bottomrule
\end{tabular}
\caption{Summary factuality measured by AlignScore.}\label{tab:alignscore}
\end{table}

%% file: misc_new/tab_oracle.tex
\begin{table}[htb]
\centering\small
\begin{tabular}{ccccc}
\toprule
\textbf{Claude-Ins.} & \textbf{R1-f} & \textbf{Align.} & \textbf{R1-f}    & \textbf{Align.}   \\\midrule
\textbf{}            & \multicolumn{2}{c}{\textbf{SamSum}} & \multicolumn{2}{c}{\textbf{CNN}}         \\
+\ours             & 41.6          & \textbf{88.0}       & 42.0             & \textbf{82.3}         \\
+Oracle             & \textbf{50.0} & 86.3                & \textbf{50.0}    & 78.8                  \\\midrule
\textbf{}            & \multicolumn{2}{c}{\textbf{ArXiv}}  & \multicolumn{2}{c}{\textbf{M.Bank}} \\
+\ours             & 45.2          & \textbf{60.0}       & 42.3             & \textbf{74.7}         \\
+Oracle             & \textbf{51.6} & 45.9                & \textbf{48.2}    & 56.7                  \\\bottomrule
\end{tabular}
\caption{Summary quality with oracle keyphrases.}\label{tab:oracle}
\end{table}

%% file: misc/result-kw-model.tex
\begin{table}[htb]
\centering\small
\setlength{\tabcolsep}{3pt}
\begin{tabular}{lccccc}
\toprule
\textbf{}                  & \textbf{SAMSum}    & \textbf{CNN}       & \textbf{M.Bank} & \textbf{Arxiv}     \\
\textbf{Method}            & \textbf{R@15} & \textbf{R@15} & \textbf{R@15}   & \textbf{R@35} \\\midrule
Rake                       & 68.3               & 11.9               & 17.1                 & 14.2               \\
TextRank                   & 80.5               & 20.8               & 19.3                 & 22.4               \\\cmidrule(lr){1-1}
\gpours                    & 75.5               & 27.7               & 40.3                 & 31.7               \\
(+32 ex.)      & 81.5               & 29.7               & 47.3                 & 32.0               \\
(+128 ex.)     & \textbf{85.5}               & 32.9               & 62.2                 & 32.7               \\\cmidrule(lr){1-1}
\ours (1k ex.)      & 83.3               & \textbf{33.6}               & \textbf{65.7}                 & \textbf{35.4}              \\\bottomrule

\end{tabular}
\caption{Keyphrase extraction performance.}\label{tab:keyword}
\end{table}

%% file: 5_con.tex
\section{Related Work}
Leveraging keyword in abstractive summarization has been explored in many works. Switching Generator-Pointer~\citep{nallapati-etal-2016-abstractive} and CopyNet~\citep{gu-etal-2016-incorporating} modify a recurrent neural network model~\citep{chopra-etal-2016-abstractive} to directly copy keywords from the source text. More recent work has adopted transformer architectures~\citep{vaswani2017attention}, which have become dominant in natural language processing. \citet{liu2022key} introduces a bias in the attention matrix to help transformer models focus on keywords. All these models need to be trained or finetuned on large-scale training data. While finetuned models typically achieve higher ROUGE scores than prompting a pretrained model, prompt-based summarizers are preferred in some industrial use cases due to their flexibility and reduced need for data collection. Incorporating keyphrases in the prompt can effectively control the length and content coverage of the summary, something that fine-tuning methods cannot easily achieve. Therefore, we cannot compare with these methods using metrics like ROUGE. 

Instruction finetuned LLMs~\citep{chung2022scaling, touvron2023llama, zhang2022opt} have shown strong performance on summarization purely via prompts, without finetuning data. Such models are often offered via APIs, enabling easier development and deployment of summarization applications. Keyphrases are still helpful for these large models, as \citet{li2023-stimulus-prompt} show that a keyphrase generater trained with reinforcement learning can improve summarization performance.

There has been interest in 2-stage extractive-then-abstractive approaches \citep{su2020two, liu2021keyword, li2021ease, su2022extract, yang2023apidocbooster}. These first extract keyphrases or sentences before abstractively summarizing them. These methods are trained end-to-end for domain-specific use cases, while our method can be pre-trained for general purpose zero-shot use cases. Practically, any keyword extractor, for example KeyBERT or LLMBERT~\cite{grootendorst2020keybert}, can be used for the first stage to enhance the summarization in the second stage. The 2-stage methods could also be implemented as Chain-of-Thought (CoT) by generating intermediate hints and final results in the same prompt, such as \citet{adams-etal-2023-sparse}. In our experiments, we compare our method with a 2-stage prompting approach -- first generating keywords using one prompt, then using those keywords for summarization in the second prompt. While slightly different from previous work, the 2-stage baseline effectively captures the use of intermediate reasoning steps of LLMs. 

\section{Conclusion}

In this paper, we propose a lightweight approach to incorporate keyphrases into the prompt for LLM-based abstractive summarization. \ours involves training a phrase extractor using supervised learning to identify salient keyphrases from the input text. These keyphrases are then injected into the prompt provided to the LLM for summary generation. We demonstrate that this approach can effectively improve the ROUGE scores of the generated summaries, indicating a higher similarity to reference summaries. Introducing keyphrases in the prompt enhances the faithfulness of the summary by ensuring that important information is captured. Additionally, our approach offers control over the length and precision/recall trade-off of the summary. Notably, our pretrained keyphrase extractor -- \gpours -- can improve summarization performance out-of-the-box without any finetuning, even in cases where training data is not available.

\section*{Limitations}

\textbf{Model Design:} We use Longformer as the backbone model to build \ours because it is light-weight and supports long context length. However, we do not evaluate the impact of using other similar-sized pre-trained language models. Additionally, we extract training labels using a fuzzy matching approach to make the model more generalizable, but more domain-specific approaches for keyphrase extraction may yield better performance.

\noindent\textbf{Evaluation:} As is common in summarization research, we rely primarily on automatic metrics and qualitative example checks to evaluate performance. These techniques have known limitations in assessing summary quality. Meanwhile, human evaluation has its own challenges. Therefore, how to best evaluate the quality of abstractive summarization models remain as an open question.


%% file: 6_appendix.tex
\onecolumn
\section*{Appendix}
\section{All Prompts}\label{sec:prompt}
Here we show all the prompts we used in the experiments. In prompt, \texttt{<text>} will be replaced with source documents, and \texttt{<keywords>} will be replaced with comma separated keyphrases extracted by \ours. We conduct light prompt engineering to get a reasonably good zero-shot prompt. 

\subsection{Zero-shot Claude Instant Prompts}
\noindent\textbf{SAMSum}
\begin{lstlisting}
Here is a conversation:
<text>

Please write a very short 1 sentence summary.
\end{lstlisting}

\noindent\textbf{SAMSum with \ours}
\begin{lstlisting}
Here is a conversation:
<text>

Please write a very short 1 sentence summary. Consider include the following information: <keywords>
\end{lstlisting}

\noindent\textbf{CNN/DailyMail}
\begin{lstlisting}
Here is a news article:
<text>

Please write a summary for the article in 2-3 sentences.
\end{lstlisting}

\noindent\textbf{CNN/DailyMail with \ours}
\begin{lstlisting}
Here is a news article:
<text>

Please write a summary for the article in 2-3 sentences. Consider include the following information: <keywords>.
\end{lstlisting}

\noindent\textbf{ArXiv}
\begin{lstlisting}
Here is a research paper:
<text>

Please write a comprehensive paper abstract section. 
\end{lstlisting}

\noindent\textbf{ArXiv with \ours}
\begin{lstlisting}
Here is a research paper:
<text>

Please write a comprehensive paper abstract section. Consider include the following information: <keywords>
\end{lstlisting}

\noindent\textbf{MeetingBank}
\begin{lstlisting}
Here is a conversation:
<text>

Please write a summary in about 5 sentences. 
\end{lstlisting}

\noindent\textbf{MeetingBank with \ours}
\begin{lstlisting}
Here is a conversation:
<text>

Please write a summary in about 5 sentences. Consider include the following information: <keywords>
\end{lstlisting}

\subsection{Zero-shot Mistral Prompts}

\noindent\textbf{SAMSum}
\begin{lstlisting}
<s>[INST]Here is a conversation:
<text>

Please write a short 1 sentence summary. [/INST]
\end{lstlisting}

\noindent\textbf{SAMSum with \ours}
\begin{lstlisting}
<s>[INST]Here is a conversation:
<text>

Please write a short 1 sentence summary. Consider include the following information: <keywords>[/INST]
\end{lstlisting}

\noindent\textbf{CNN/DailyMail}
\begin{lstlisting}
<s>[INST]Here is a news article:
<text>

Please write a short summary for the article in 1-2 sentences.[/INST]
\end{lstlisting}

\noindent\textbf{CNN/DailyMail with \ours}
\begin{lstlisting}
<s>[INST]Here is a news article:
<text>

Please write a short summary for the article in 1-2 sentences. Consider include the following information: <keywords>[/INST]
\end{lstlisting}

\noindent\textbf{ArXiv}
\begin{lstlisting}
<s>[INST]Here is a research paper:
<text>

Please write a short abstract in about 3 sentences.[/INST]
\end{lstlisting}

\noindent\textbf{ArXiv with \ours}
\begin{lstlisting}
<s>[INST]Here is a research paper:
<text>

Please write a short abstract in about 3 sentences. Consider include the following information: <keywords>[/INST]
\end{lstlisting}

\noindent\textbf{MeetingBank}
\begin{lstlisting}
<s>[INST]Here is a conversation:
<text>

Please write a 2-3 sentence summary.[/INST]
\end{lstlisting}

\noindent\textbf{MeetingBank with \ours}
\begin{lstlisting}
<s>[INST]Here is a conversation:
<text>

Please write a 2-3 sentence summary. Consider include the following information: <keywords>[/INST]
\end{lstlisting}

\subsection{Zero-shot Falcon and Flan-T5 Prompts}
\noindent\textbf{SAMSum}
\begin{lstlisting}
Here is a conversation:
<text>

Please write a short 1 sentence summary. Summary: 
\end{lstlisting}

\noindent\textbf{SAMSum with \ours}
\begin{lstlisting}
Here is a conversation:
<text>

Please write a short 1 sentence summary. Consider include the following information: <keywords>

Summary:
\end{lstlisting}

\noindent\textbf{CNN/DailyMail}
\begin{lstlisting}
Here is a news article:
<text>

Please write a short summary for the article in 1-2 sentences.

Make sure the summary is no more than 2 sentences. Summary: 
\end{lstlisting}

\noindent\textbf{CNN/DailyMail with \ours}
\begin{lstlisting}
Here is a news article:
<text>

Please write a short summary for the article in 1-2 sentences. Consider include the following information: <keywords>.

Make sure the summary is no more than 2 sentences. Summary: 
\end{lstlisting}

\noindent\textbf{ArXiv}
\begin{lstlisting}
Here is a research paper:
<text>

Please write a short abstract in about 3 sentences.

Abstract: 
\end{lstlisting}

\noindent\textbf{ArXiv with \ours}
\begin{lstlisting}
Here is a research paper:
<text>

Please write a short abstract in about 3 sentences. Consider include the following information: <keywords>.

Abstract: 
\end{lstlisting}

\noindent\textbf{MeetingBank}
\begin{lstlisting}
Here is a conversation:
<text>

Please write a 2-3 sentence summary.

Summary: 
\end{lstlisting}

\noindent\textbf{MeetingBank with \ours}
\begin{lstlisting}
Here is a conversation:
<text>

Please write a 2-3 sentence summary. Consider include the following information: <keywords>

Summary: 
\end{lstlisting}

\section{Loss Function}
The training objective for \ours is 
\[
\mathcal{L}  = -\sum_{i \in 1\ldots T} \sum_{k\in l_i \ldots r_i} [\text{label}(p_i) \log f(x_k)  + \lambda (1-\text{label}(p_i)) \log (1 - f(x_k))\big],
\]
where $f(x_k)$ denotes the binary classification probability on token $x_k$ by the classification head, and $\lambda$ is the class balancing weight. 

\clearpage

\input{misc/failed_exp_mistral}

\input{misc/case-study}

\input{misc/tab_granularity_vis}

\input{misc_new/tab_dataset}
\clearpage

%% file: misc/failed_exp_mistral.tex
\begin{table*}[tb]
\centering\small
\begin{tabular}{rp{12cm}}
\toprule
\textbf{Document} & \begin{tabular}[c]{@{}p{12cm}@{}}Jake: \myhl{we don't have english today} \\ Nadia: whooooah! \\ Vanessa: cool! you sure though? \\ Jake: Smith called in sick, they couldn't find a replacement, oh yeah\end{tabular} \\
\textbf{Keyphrases} & Vanessa; Smith called; \myhl{Jake; replacement;} english today \\
\textbf{Mistral} & Jake announced that they don't have English class today due to Smith being sick and unable to find a replacement. \\
\textbf{+\ours} & Vanessa is excited that \myhl{Jake is teaching English today} instead of Smith, who called in sick and couldn't be replaced \\
\midrule
\textbf{Document} & \begin{tabular}[c]{@{}p{12cm}@{}}Sam: Good afternoon, is this available? \\ Lia: Hi, how can I help you? \\ Sam: I'm interested in this product. Are the earphones available? \\ Lia: Yes, they are. \\ Sam: where can I collect them? \\ Lia: Well, you can have them delivered to you through courrier - this costs £4 (pay on delivery) \\ Lia: I'm afraid collecting them personally is a little difficult, as the product can only be collected in Garstang (where the company is located) \\ Sam: Okay. \\ Sam: Are they new? Do you have any other colors that white? \\ Lia: Completely new - we also offer grey and pink. \\ Sam: Then I would like the white ones please. \\ Lia: Okay, so I just need your address and telephone \\ Sam: The total costs £187 - can I lower the price to £180?? \\ Lia: \myhl{fraid not - company policy.} \\ Sam: Oh, okay. \\ Sam: How long does it take for the product to arrive? \\ Lia: It generally arrives the next day, in rare instances it's longer.\end{tabular} \\
\textbf{Keyphrases} & Lia; earphones available; \myhl{Garstang; £180; collected;} Sam; product; white; delivered; company. \\
\textbf{Mistral} & Sam is interested in purchasing earphones and asks about their availability, collection options, and delivery time. \\
\textbf{+\ours} & Lia confirmed that the earphones are available, can be delivered for £4, and are only available for collection in Garstang, but offered Sam the \myhl{option to purchase them for £180} \\
\midrule
\textbf{Document} & \begin{tabular}[c]{@{}p{12cm}@{}}Kim: I'm going to Seoul! \\ Agatha: wow finally! \\ Mark: When? Bring us some kimchi! \\ Kim: \myhl{People on a plane will kill me if I do} :D \\ Kim: I'm going in April <unk>3 \\ Agatha: Are you going with Jane? \\ Kim: yes, of course - she'll be our tour guide \\ Mark: so jealous... I'd love to come as well \\ Kim: Really? I asked you like a hundred times\end{tabular} \\
\textbf{Keyphrases} & Seoul; Kim; Jane; Agatha; Mark; going; \myhl{kimchi;} tour guide; Kim. \\
\textbf{Mistral} & Kim is going to Seoul in April with Jane as their tour guide and Agatha and Mark express their excitement and jealousy. \\
\textbf{+\ours} & Kim is going to Seoul in April with Jane as their tour guide, and \myhl{they will bring some kimchi} back for Mark who is jealous and wishes he could come too \\
\bottomrule
\end{tabular}
\caption{Examples where adding keyphrases in prompt hurts the faithfulness of the summary. The failure pattern is if a keyphrase is negated in the document, the LLM would ignore the negation.}\label{tab:mistral-faith}
\end{table*}

%% file: misc/case-study.tex
\begin{table*}[t]
\centering\small
\begin{tabular}{lp{12cm}}
\toprule
\textbf{Gold Summary}      & NHS gave out 404,500 prescriptions for suncream at a cost of £13m in 2014. Also handed out 4.7million prescriptions for indigestion pills costing £29m. Other items routinely prescribed include vitamins, Vaseline and toothpaste. Critics branded prescriptions 'ludicrous' at time of financial crisis for NHS.                                                                                                              \\
\textbf{Key Phrases} & toothpaste; suncream; patients; \myhl{404,500 prescriptions}; NHS; \myhl{indigestion pills costing£29million}; NHS handed; \myhl{£13million}; spending millions; prescribed                                                                                                                                                                                                                                                                          \\
\textbf{Mistral-7B} & The NHS is spending millions on prescriptions for suncream and other household essentials, according to a damning analysis. The figures come as the UK ranks below many European countries in health and wellness, and as the NHS faces a huge financial crisis.                                                                                                                                                                \\
\textbf{+\ours}     & The NHS is spending millions of pounds on prescribing \myhl{toothpaste} and suncream to patients, according to a damning analysis. In 2014, the NHS handed out \myhl{404,500 prescriptions} for suncream at a cost of \myhl{£13million} and 4.7million prescriptions for \myhl{indigestion pills costing£29million}.                                                                                                                                        \\\midrule
\textbf{Gold Summary}      & Cloned camel was born in 2009 and called Inzaz, which means achievement. She was cloned from ovarian cells and born by surrogate mother. Injaz is now six years old and is said to have conceived naturally. She's expected to give birth late this year, proving cloned animals' fertility.                                                                                                                                    \\
\textbf{Key Phrases} & Injaz; pregnant; slaughtered camel; cloned; ovarian cells; surrogate mother; first cloned camel; \myhl{2009}; \myhl{conceived naturally}; whose name                                                                                                                                                                                                                                                                                          \\
\textbf{Mistral-7B} & The world’s first cloned camel, Injaz, is pregnant and expected to have a calf. This marks the first time that a cloned camel has been able to reproduce naturally, and scientists are excited to see if cloned camels are as fertile as naturally produced ones.                                                                                                                                                               \\
\textbf{+\ours}     & Injaz, the world's first cloned camel born from the ovarian cells of a slaughtered camel in \myhl{2009}, is now pregnant and expected to give birth to a calf later this year. The camel's \myhl{conception was natural}, proving that cloned camels are fertile and capable of reproducing.                                                                                                                                                  \\\midrule
\textbf{Gold Summary}      & Lewis Hamilton won Sunday's Barhain Grand Prix ahead of Kimi Raikkonen. Hamilton is out of contract at the end of the year and is yet to sign new deal. It has been suggested that Hamilton could replace Raikkonen at Ferrari. But team principal Maurizio Arrivabene says he is happy with driver line-up.                                                                                                                    \\
\textbf{Key Phrases} & Kimi Raikkonen; Ferrari; \myhl{next season}; \myhl{replace Raikkonen}; Mercedes; 30; Lewis Hamilton yet; Bahrain; Ferrari team principal Maurizio Arrivabene; Vettel                                                                                                                                                                                                                                                                          \\
\textbf{Mistral-7B} & Ferrari team principal Maurizio Arrivabene has revealed that he uses a carrot-and-stick method with Kimi Raikkonen to keep his desire to remain with the Maranello marque high. The Finn has arguably never been happier during his Formula One career, thanks to the atmosphere generated within the team by the arrival of Arrivabene.                                                                                        \\
\textbf{+\ours}     & Ferrari team principal Maurizio Arrivabene has revealed that he uses a carrot-and-stick method to keep Kimi Raikkonen's desire to remain with the Maranello marque high. With Lewis Hamilton yet to sign a new contract, it has been suggested that he will \myhl{replace Raikkonen} at Ferrari for \myhl{next season}. In the Bahrain Grand Prix, Raikkonen finished on the podium for the first time in 26 races as runner-up to Hamilton.
\\\bottomrule
\end{tabular}
\caption{Examples of using \ours with Mistral-7B model on CNN dataset.}\label{tab:case}
\end{table*}

%% file: misc/tab_granularity_vis.tex
\begin{table*}[htb]
    \centering\small
    \begin{tabular}{rp{12cm}}
    \toprule
    
\textbf{Document} & \begin{tabular}[c]{@{}p{12cm}@{}}\myhlb{Jenkin} : hey what is your \myhlb{spirit} \myhlb{animal} ? \\\myhlb{Sophie} : what ? \\\myhlb{Jenkin} : go on ? \\\myhlb{Sophie} : I dont know a \myhlb{fox} lol \\\myhlb{Jenkin} : are you wiley ? \\\myhlb{Sophie} : sometimes \\\myhlb{Jenkin} : I am a \\\myhlb{Sophie} : I think you are a bit mad like the mad \\\myhlb{Jenkin} : I have been \myhlb{reading} about \myhlb{animal} \myhlb{spirits} its quite good \\\myhlb{Sophie} : you will have to tell me about the \myhlb{fox} .. do you decide what your \myhlb{animal} is or does someone tell you ? \\\myhlb{Jenkin} : There is a \myhlb{pack} of \myhlb{cards} and you \myhlb{choose} the one that you are \myhlb{drawn} to \\\myhlb{Sophie} : oh right I \myhlb{would} \myhlb{choose} the \myhlb{Fox} \\\myhlb{Jenkin} : well I did n't know but I was \myhlb{drawn} to the \myhlb{dolphin} \\\myhlb{Sophie} : oh \\\myhlb{Jenkin} : I will \myhlb{bring} them over \myhlb{tomorrow} \\\myhlb{Sophie} : oh yes please that will be great  \end{tabular} \\
\textbf{Reference} & Jenkin has been reading about spirit animals and he was drawn to a dolphin. Sophie would choose a fox. Jenkin will bring pack of cards with spirit animals to Sophie tomorrow. \\
\midrule
\textbf{Document} & 
\begin{tabular}[c]{@{}p{12cm}@{}}\myhlb{Jacky} : I think you were \myhlb{right} yesterday . \\\myhlb{David} : What about ? I 'm \myhlb{right} about most \myhlb{things} : P \\\myhlb{Jacky} : Yeah , whole you ; ) \\\myhlb{Jacky} : About \myhlb{taking} the \myhlb{blame} etc . \\\myhlb{David} : Okey , I remeber . We 'll \myhlb{talk} later ? \\\myhlb{Jacky} : With pleasure . I 'll call you when I get \myhlb{home} . \end{tabular}\\
\textbf{Reference} & According to Jacky, David did the right thing taking the blame. They will talk when Jack comes back home. \\
\midrule
\textbf{Document} & 
\begin{tabular}[c]{@{}p{12cm}@{}}\myhlb{Jill} : So \myhlb{bored} ! \\\myhlb{Nate} : Well ... ca n't help you there \\\myhlb{Nate} : Still at \myhlb{work} \\\myhlb{Jill} : ugh I need to find a job \\\myhlb{Jill} : I 've \myhlb{watched} everything on \myhlb{youtube} already \\\myhlb{Nate} : Doubt it : P I 'll \myhlb{call} you when I get off \myhlb{work}  \end{tabular}\\
\textbf{Reference} & Jill is bored and has watched YouTube. Nate is at work and will call Jill when he finishes it.\\
\bottomrule
\end{tabular}
\caption{Visualization of overlapping words between the document and reference summary on the SAMSum dataset. The words are dispersed across the document, making it difficult to extract sentence-level salient information.}
\label{tab:gran_vis}
\end{table*}

%% file: misc_new/tab_dataset.tex
\begin{table}[htb]
\centering\small
\begin{tabular}{llc}
\toprule
\textbf{Dataset} & \textbf{Description}                                    & \textbf{Input/Output} \\\midrule
CNN    & News article headline generation                        & 773/58              \\
SAMSum           & Messenger-like conversations summarization      & 127/23              \\
ArXiv            & Research paper abstract generation                      & 6446/166             \\
MeetingBank      & Meeting transcript summarization & 3095/66             \\\bottomrule
\end{tabular}
\caption{Dataset description and input/output length.}\label{tab:dataset}
\end{table}